\documentclass[11pt]{article}

\newcommand{\mdiv}{\nabla\cdot}
\newcommand{\mvec}[1]{{\bf #1}}
\newcommand{\mhvec}[1]{{\bm{\hat #1}}}
\newcommand{\mcurl}{\nabla\times}
\newcommand{\nabb}{\nabla^2}

\newcommand{\theosea}{{\em TheoSea}}
\newcommand{\alert}[1]{{\color{red}\bf ALERT! #1}}
\newcommand{\impl}[1]{{\color{green}\bf IMPLEMENT! #1}}
\newcommand{\cmplx}{{\cal M}}
\newcommand{\fld}{{\cal F}}

\usepackage{graphicx}
\usepackage{color}
\usepackage{amsmath}
\usepackage{bm}

\title{On the enumeration of sentences by compactness}

\author{Mark A.~Stalzer \\
  Center for Data-Driven Discovery \\
  California Insitute of Technology \\
  stalzer at caltech.edu \\ \\
  A talk on this paper will be presented at Juliacon 2017, Berkeley, CA}

\begin{document}

\maketitle

\begin{abstract}
Presented is a Julia meta-program that discovers compact theories from data if they exist. It writes candidate theories in Julia and then validates: tossing the bad theories and keeping the good theories. {\em Compactness} is measured by a metric: such as the number of space-time derivatives. The underlying algorithm is applicable to a wide variety of combinatorics problems and compactness serves to cut down the search space.
\end{abstract}

\section{Introduction}
 
 This work flowed from a comment in the concluding remarks of a recent review (2016) of work in data-driven scientific discovery\cite{stalzer16}. Specifically,
\begin{quote}
  {\em \ldots it may be within current computing and algorithmic technology to infer the Maxwell Equations directly from data given knowledge of vector calculus.}
\end{quote}
This short paper reports on recent progress towards this objective in the broader context of combinatorics and there are applications beyond science. The overarching goal is to develop methods that can quickly discover compact theories from data. Most data intensive analysis techniques are based on machine learning or statistics. They are quite useful, but do not lead to deep understanding or insight.

Abstractly, think of an alphabet ${\cal A} = [A, B, C, \ldots]$ where any letter can appear once in a sentence and the length of the alphabet is $n.$ This is a simple combinatorial enumeration problem and the solution to the number of sets of size $m$ (later $m$ will be relabeled $q$) taken from ${\cal A}$ is $C(n, m).$ However, what if the symbols --- letters --- in the alphabet have different weights? What if the alphabet is more like ${\cal A} = [A=1, B=1, C=4, D=4, E=4, F=4, G=7, H=7, I=7, J=7, K=4,  L=7].$ This can dramatically decrease the enumeration size as shown in the next section, and the underlying motivation is shown in Sec.~\ref{sec:phys}.

\paragraph{Julia.} The code is written in Julia\cite{julia17}, a relatively recent language (roughly 2012) that is both easy to use and has high performance. Julia can be programmed at a high expressive level, and yet given enough type information it automatically generates efficient machine code. The code is a Julia {\em meta-program} that writes candidate theories in Julia that are then validated against data. Ironically, one optimization that is used is to circumvent the Julia compilation step when first validating a theory: if the theory looks promising it can then be compiled for further, more rapid, validation. The full code with some partial results is shown in the Appendix.

\section{Algorithm}
\label{sec:compsea}

The algorithm enumerates sets of increasing complexity $q,$ where $q$ is the sum of the alphabet letter weights in a given candiate theory. It can be thought of as a form Depth-First Iterative Deepening (DFID)\cite{korf85} first formalized by R.~E.~Korf in 1985. Optimality flows from a theorem by Korf:

\newtheorem{theo}{Theorem}

\begin{theo}[Korf 4.2]
Depth-first iterative-deepening is asymptotically optimal among brute-force tree searches in terms of time, space, and length of solution.
\end{theo}

By length of solution, Korf means the depth of the search where a solution is found. For this code compactness is the sum of the symbol weights along a potential solution branch in the search; as will be shown in Sec.~\ref{sec:phys}.

It is perhaps easiest to think of the algorithm inductively. There is a data structure {\tt theos} that holds all theorems (sets) of length $q$ and it is built up from $q = 1.$ The base cases are the singleton theories of a given complexity, so for the alphabet ${\cal A}$ we have {\tt theos[1] = [A, B]} and {\tt theos[4] = [D, ...]} and so on. So the base cases, such as $q=1$ are all set; and then for $q>1$ we use a $q: l,m$ ``Squeeze''. At step $q$ consider all theories that can possibly be of length $q,$ {\em marching} $l$ upward from 1 and $m$ downward from $q-1$ in a kind of double iteration. The correctness is immediate by Korf 4.2 and the fact that $q = l+m,$ too short theories are discarded $(<q)$, and set elements are unique. The Julia code is in the Appendix.

\begin{figure}[t]
\centering
\includegraphics[scale=0.8]{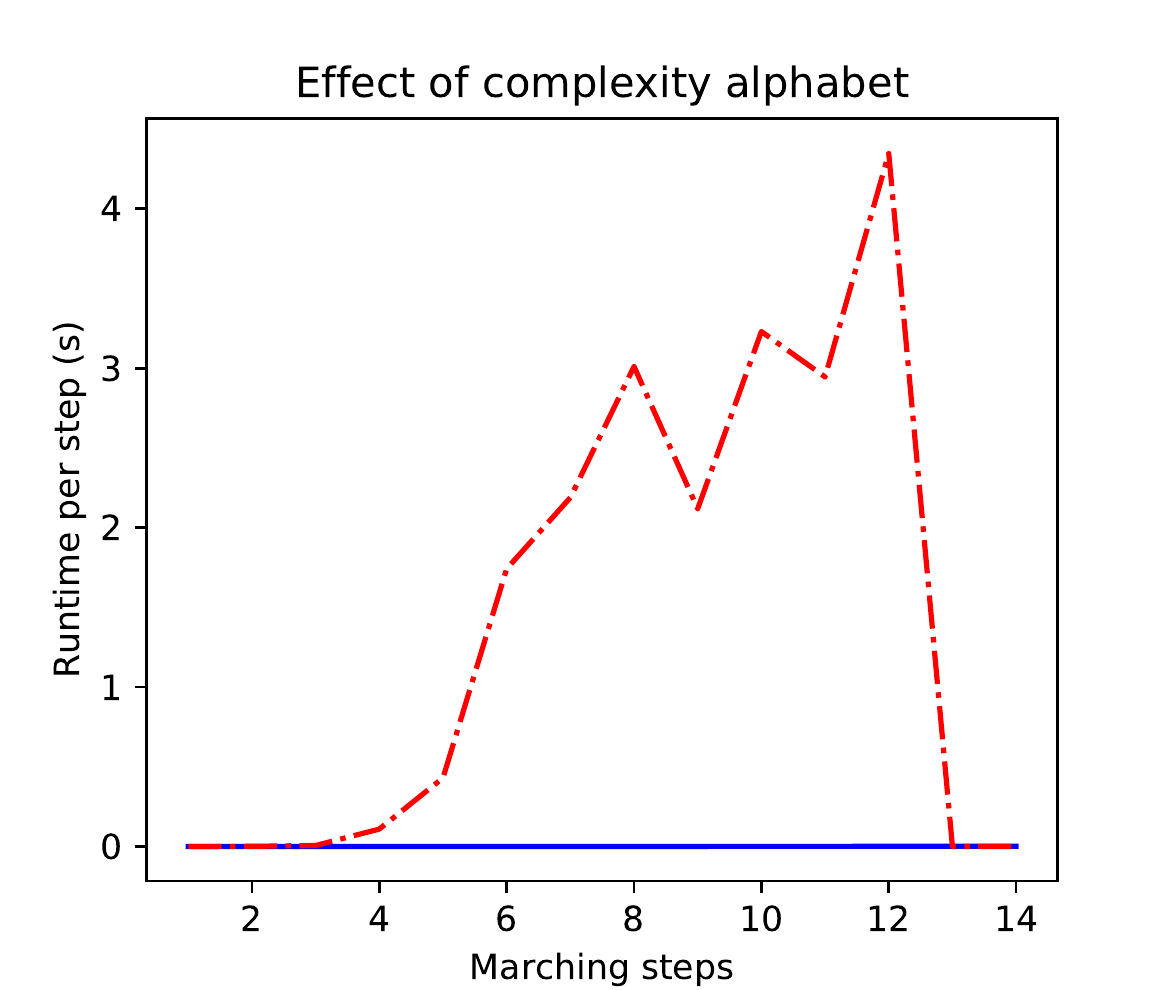}
\caption{Time to discovery (s): Fast versus Slow brute force. The singleton search cuts off at $[A, \ldots, L]$ (12) due to maximum complexity.}
\label{fig:power}
\end{figure}

\paragraph{Performance.}  The timings are in shown in Tab.~\ref{tab:data}. The total times are ${\rm Fast} =0.006s$ and ${\rm Slow} =20.1s.$\footnote{The machine was a MacBook Pro (Retina, 13-inch, Late 2013) running a 2.8 GHz Intel Core i7 single-threaded. The version of Julia was 0.5. The time varies a bit depending on my laptop's mood.} A graph is in Fig.~\ref{fig:power} -- {\em compactness matters.}

\begin{table}
\centering
\begin{tabular}{lll}
\hline
Compexity depth & Slow (s) & Fast (s) \\
\hline
1 & 4.5201e-5 & 5.2887e-5 \\
2 &  0.000665346 & 7.4057e-5 \\
3 & 0.006004517 & 3.896e-5 \\
4 &  0.108836084 & 5.9633e-5 \\
5 & 0.43004444 & 8.6931e-5 \\
6 & 1.741446458 & 0.000147389 \\
7 &  2.188128388 & 0.000136736 \\
8 &  3.010306449 & 0.000173676 \\
9 &  2.118029821 & 0.000326116 \\
10 & 3.229660437 & 0.000412358 \\
11 &  2.94480559 & 0.000377891 \\
12 & 4.34652684 & 0.000894144 \\
13 & 0. & 0.001261163 \\
14 & 0. & 0.001051088  \\
\hline
\end{tabular}
\caption{Code speed in Julia against itself without compactness.}
\label{tab:data}
\end{table}

\section{Motivation}
\label{sec:phys}
The underlying motivation was described in the Introduction, namely rediscovering the Maxwell Equations. Here is the decoder ring.

\begin{table}[h]
\centering
\begin{tabular}{lll}
\hline
Operator & Term Cost & Alphabet \\
\hline
$\fld$ & 1 & A, B \\
$\mdiv\fld$ & 4 & C, D\\
$\mcurl\fld$ & 7 & G, H\\
$\nabb\fld$ & 7 & I, J\\
 ${\partial \over \partial t}\fld$ & 4 & E, F\\
${\partial^2 \over \partial t^2}\fld$ & 7 & K, L\\
\hline
\end{tabular}
\caption{Complexity of each operator term working on a field $\fld \in \{\mvec{E}, \mvec{B}\}.$}
\label{tab:tcmplx}
\end{table}

The Maxwell Equations, up to constants, are $[C], [D],$ the $\mvec{E}, \mvec{B}$ divergence equations, both of complexity 4; and the field coupling equations of $[G, F]$ and $[H, E],$ of complexity 11. Also, the fields equations of light are $[I, K]$ and $[J, L],$ each of complexity 14\cite{fitzpatrick08}. The complexity metric is just $1 +$ the number of space-time derivatives taken.

There are many avenues for future development as briefly listed below.
\begin{itemize}
\item Validation and constant determination. Fundamental constants such as $c$ were known at the time of Maxwell. However, constants can be automatically determined via linear algebra in this case.
\item Bigger data and parallelism. The data sets needed for the Maxwell Equations re-discovery are small but semantically very rich. Other data sets, such as for macro economics, will be far larger. Here Julia's on-the-fly compilation (of candidate theories) and support for parallel processing with be very helpful, and this is one of the reasons the language was chosen\footnote{The author encourages the Julia developers to continue work on threads as that model is natural for many-core processors. For example, the main thread could enumerate candidate theories and then send them to several worker threads for validation. At any instant several theories would be under consideration.}.
\item The fully general equations can be re-discovered just by adding a current $\mvec{J}$ and source region $\rho$ to the alphabet ${\cal A}$ with the appropriate validation step.
\end{itemize}
But, perhaps the most exciting extension is to apply this search algorithm to other domains such as thermodynamics, macro economics, and chaos. Work is progressing in these areas, and focusing on the applicable representation language and semantics.

The purpose of this short paper was not to get into the motivating physics, but to look at the combinatorics which is of much broader utility.

\section*{Materials}

The Julia code is attached in the Appendix. The code is distributed under a Creative Commons Attribution 4.0 International Public License.

\section*{Acknowledgements}

This research is funded by the Gordon and Betty Moore Foundation through Grant GBMF4915 to the Caltech Center for Data-Driven Discovery. Discussions with Mr.~William Xu of Caltech Math/Computer Science were very helpful. The author is grateful to Prof.~S.G.~Djorgovski of Caltech Astronomy and Prof.~V.~Chandler of KGI Natural Sciences for their support.

\section*{Appendix}
\label{sec:code}
\begin{verbatim}
julia> theos
14-element Array{Array{Set{Char},N},1}:
 Set{Char}[Set(['A']),Set(['B'])]                                                                                                                                                                                                                                                                                                                                                                                                                                                          
 Set{Char}[Set(['A','B'])]                                                                                                                                                                                                                                                                                                                                                                                                                                                                 
 Set{Char}[]                                                                                                                                                                                                                                                                                                                                                                                                                                                                               
 Set{Char}[Set(['C']),Set(['D']),Set(['E']),Set(['F'])]                                                                                                                                                                                                                                                                                                                                                                                                                                    
 Set{Char}[Set(['A','C']),Set(['B','C']),Set(['A','D']),Set(['B','D']),Set(['E','A']),
   Set(['E','B']),Set(['A','F']),Set(['B','F'])]                                                                                                                                                                                                                                                                                                                                                        
 Set{Char}[Set(['A','B','C']),Set(['A','B','D']),Set(['E','A','B']),Set(['A','B','F'])]                                                                                                                                                                                                                                                                                                                                                                                                    
 Set{Char}[Set(['G']),Set(['H']),Set(['I']),Set(['J']),Set(['K']),Set(['L'])]                                                                                                                                                                                                                                                                                                                                                                                                              
# ... up to complexity 14.

#-*- mode: Julia
# TheoSea enumeration algorithm, May 29, 2017
#
# Mark A. Stalzer, Caltech, stalzer@caltech.edu
#
# The code is distributed under a Creative Commons Attribution 4.0
# International Public License. If you use this work please attribute
# to M. Stalzer, "On the enumeration of sentences by compactness'',
# arXiv, June 2017.
#
# This research was funded by the Gordon and Betty Moore Foundation
# through Grant GBMF4915 to the Caltech Center for Data-Driven Discovery.

# Complexity alphabet, a simple one
# alph = [('A', 1), ('B', 1), ('C', 3), ('D', 4), ('E', 4)]

# Contrast these two alphabets with the same number of symbols to
# see how much complexity helps in the enumeration time.
alph = [('A', 1), ('B', 1), ('C', 4), ('D', 4), ('E', 4),
  ('F', 4), ('G', 7), ('H', 7), ('I', 7), ('J', 7), ('K', 7), ('L', 7)]
#
#alph = [('A', 1), ('B', 1), ('C', 1), ('D', 1), ('E', 1),
#  ('F', 1), ('G', 1), ('H', 1), ('I', 1), ('J', 1), ('K', 1), ('L', 1)]
#

# Maximal possible theory size q with user limit MAX_COMP
MAX_COMP = 14
max_q = min(sum([a[2] for a in alph]), MAX_COMP)

# Seed theos data structure with singleton theories by complexity
theos = Array{Array{Set{Char}}}(max_q)

function init_theos()
    fill!(theos, [])

    for a in alph
        if theos[a[2]] == [] theos[a[2]] = [Set([a[1]])]
        else push!(theos[a[2]], Set([a[1]]))
        end
    end

    global valids = []
end

# Theory validation function for candidate theories
function valid(cand_theo)
    true
end

# March theories of increasing complexity q until max_q
function march() for q in 1:max_q
    tic()    
    # q:l,m Squeeze
    #
    # All singleton theories of size q already in theos as base cases
    # Generate other theories by combinations of l+m=q squeezing together
    m = q-1; l = 1
    while m >= l
        for m_el in theos[m]
            for l_el in theos[l]
                ml_el = union(m_el, l_el)
                if length(ml_el) == length(m_el) + length(l_el) # < q?
                    if theos[q] == [] theos[q] = [ml_el]
                    elseif !(ml_el in theos[q]) # Nonredundant
                        push!(theos[q], ml_el)
                    end  
                end
            end
        end
        m -= 1; l += 1
    end

    # VALIDATE final theos[q]
    append!(valids,
      [(q, i) for i in 1:length(theos[q]) if valid(theos[q][i])])
    println(toq())
end end

# init_theos, march once, init_theos, march twice to get timings right
init_theos()
@time march()
init_theos()
@time march()
\end{verbatim}

% References

\end{document}